\documentclass[runningheads]{llncs}

\usepackage{cite}
\usepackage{graphicx}
\usepackage{tikz}
\usepackage{tabularx}
\usepackage{booktabs}
\usepackage{subfig}
\usepackage{url}
\usepackage{hyperref}

\begin{document}
\title{The BARN Challenge 2023 - Autonomous Navigation in Highly Constrained Spaces Inventec Team}
\titlerunning{Autonomous Navigation in Highly Constrained Spaces - Inventec Team}

\author{Hanjaya Mandala\inst{1} \and Guilherme Christmann\inst{1}}
\institute{AI Center, Inventec Corporation, Taipei, Taiwan
\email{\{hsu.hanjaya,guilherme.christmann\}@inventec.com}
}

\maketitle

\begin{abstract}
Navigation in the real-world is hard and filled with complex scenarios. The Benchmark Autonomous Robot Navigation (BARN) Challenge is a competition that focuses on highly constrained spaces. Teams compete using a standard platform in a simulation and a real-world stage, with scenarios ranging from easy to challenging. This technical report presents the system and methods employed by the Inventec Team during the BARN Challenge 2023 \footnote{\url{https://cs.gmu.edu/~xiao/Research/BARN_Challenge/BARN_Challenge23.html}}. At its  core, our method uses the baseline learning-based controller LfLH \cite{wang2021agile}. We developed extensions using a finite state machine to trigger recovery behaviors, and introduced two alternatives for forward safety collision checks, based on footprint inflation and model-predictive control. Moreover, we also present a backtrack safety check based on costmap region-of-interest. Compared to the original baseline, we managed a significant increase in the navigation score, from 0.2334 to 0.2445 (4.76\%). Overall, our team ranked second place both in simulation and in the real-world stage. Our code is publicly available at: \url{https://github.com/inventec-ai-center/inventec-team-barn-challenge-2023.git}
\keywords{Navigation  \and State machine \and Recovery behavior.}
\end{abstract}

\section{Introduction}
\par In the real-world, environments are often cluttered and primarily designed for humans with consideration for robots in second. Navigating in the real-world is hard and filled with complex scenarios. As such, the development of capable navigation systems is imperative \cite{nahavandi2022comprehensive}. Moreover, navigation robots are also required to move safely and efficiently. Navigation in highly constrained spaces \cite{perille2020benchmarking} and dynamic environments encompasses a wide range of robot navigation applications such as warehouse automation, search and rescue operations, and household assistance.

\begin{figure}
    \centering
    \includegraphics[width=0.75\columnwidth]{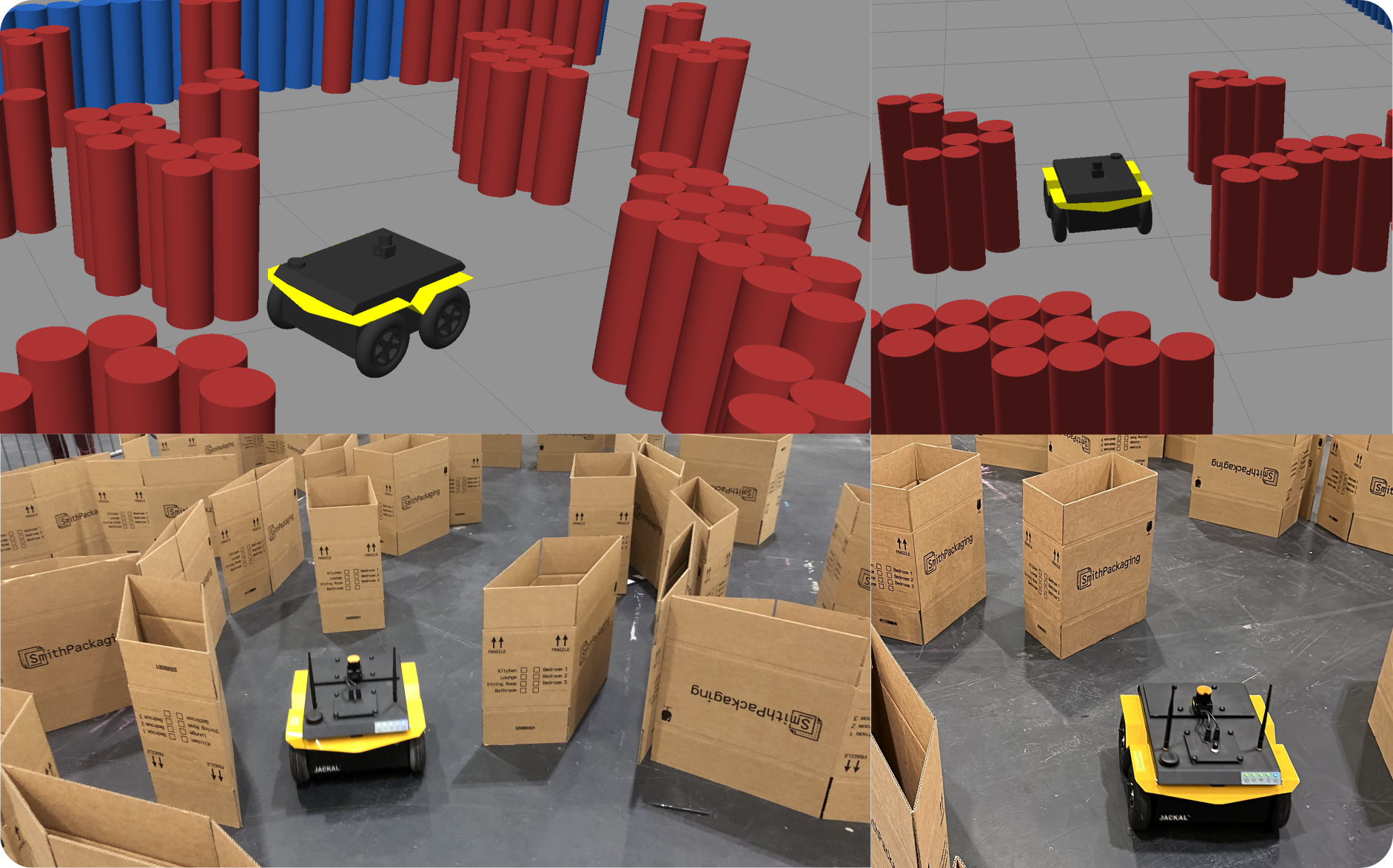}
    \caption{BARN Challenge in simulation and real-world environments.}
    \label{fig:teaser}
\end{figure}

\par A robust navigation robot is required to be able to find the best route \cite{gasparetto2015path, zhang2018path}, safely navigate through narrow passages \cite{kamil2015review, pandey2017mobile}, avoid obstacles \cite{pandey2017mobile}, maneuver precisely \cite{medvedev2021path}, and perform self-recovery under any circumstances \cite{yuan2018lost}. Navigation is a widely researched topic in the literature \cite{gul2019comprehensive}. Researchers have explored classical path-planning algorithms such as the widely used DWA \cite{fox1997dynamic}, E-Band \cite{quinlan1993elastic}, and TEB\cite{rosmann2017kinodynamic}. Machine learning based methods have also become popular in robot navigation \cite{xiao2022motion}, including supervised learning \cite{xiao2021toward}, self-supervised learning \cite{wang2021agile}, and reinforcement learning algorithms \cite{zhu2021deep}.

\par To benchmark the progress of navigation in complex environments, the BARN Challenge was created \cite{perille2020benchmarking}. It promotes research in state-of-the-art controllers that handle cluttered and novel scenarios, critical for real-world deployment. This technical report presents the Inventec Team's system and methods utilized in the BARN Challenge 2023. First, we provide an overview of the BARN competition. Then, we present our technical methodology. Next, we elaborate on our results and discuss the lesson that we learned. Finally, we conclude and mention our future work.

\section{Competition}    
    \par The BARN Challenge is a yearly competition \cite{xiao2022autonomous} held during ICRA since 2022 that aims to develop a state-of-the-art controller to navigate in highly constrained spaces without collisions. However, setting up evaluations in the real-world is difficult, time-consuming, and expensive. To lower the barrier of entry, the first stage of the challenge is done in simulation. The top-performing teams are invited to the real-world stage, where a standard platform is provided to all participants.
    
    \par In this section we introduce the robot platform, the competition scenarios, and how the score is evaluated.

    \subsection{Standard Platform - Jackal Robot}        
        \par
\begin{figure}
            \centering
            \includegraphics[trim={2cm 2cm 2.5cm 2cm},clip=true, width=0.45\columnwidth]{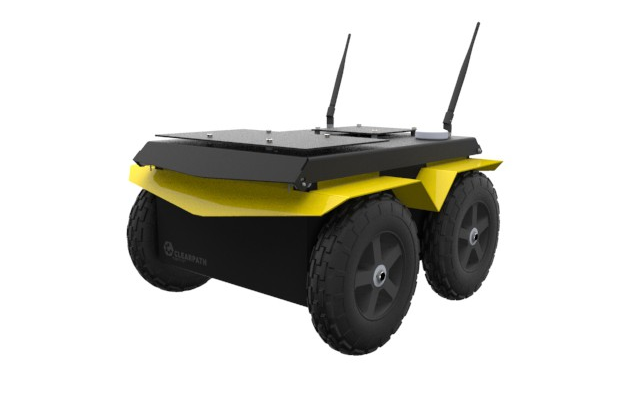}
            \caption{The Clearpath Jackal robot.}
            \label{fig::jackal_robot}
        \end{figure}

 The BARN challenge utilized the four-wheeled differential robot Jackal from Clearpath Robotics, shown in Fig. \ref{fig::jackal_robot}. The robot is equipped with 2D Hokuyo UST-10LX LiDAR, which provides a 270-degree field-of-view (Fig. \ref{fig::jackal_lidar_costmap}). Despite the four-wheeled design of the Jackal, it uses only two motors that connect the front and rear wheels on each side. The robot can be controlled via ROS. In the simulation phase, the robot configuration and control are replicated in the Gazebo simulation environment.

    \subsection{Scenarios}
\begin{figure}[t]
            \centering
            \subfloat[\centering World Index 66]{{\includegraphics[trim={15cm 5cm 15cm 0cm},clip=true, width=5cm]{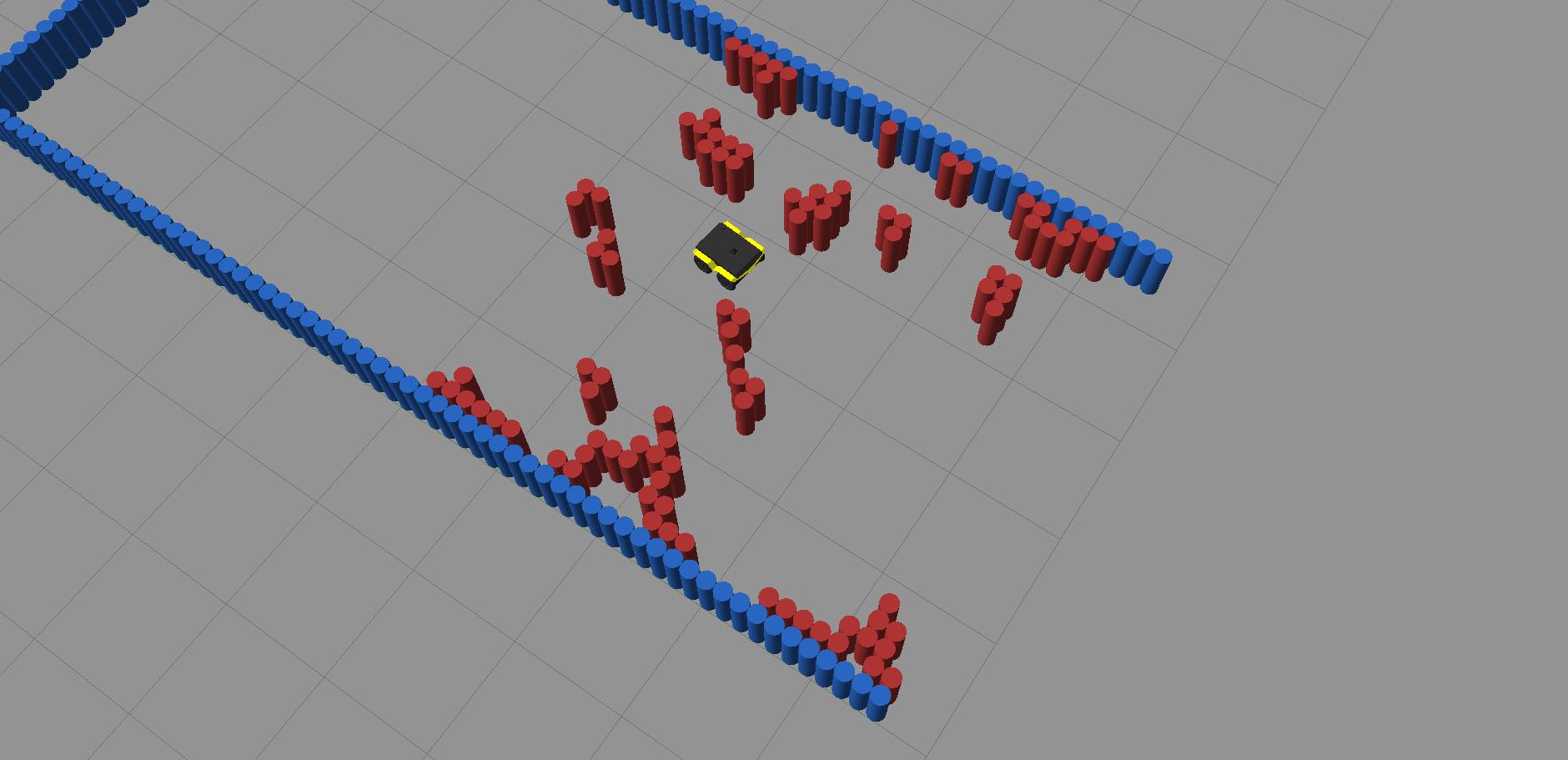} }}%
            \quad
            \subfloat[\centering World Index 114]{{\includegraphics[trim={15cm 5cm 15cm 0cm},clip=true, width=5cm]{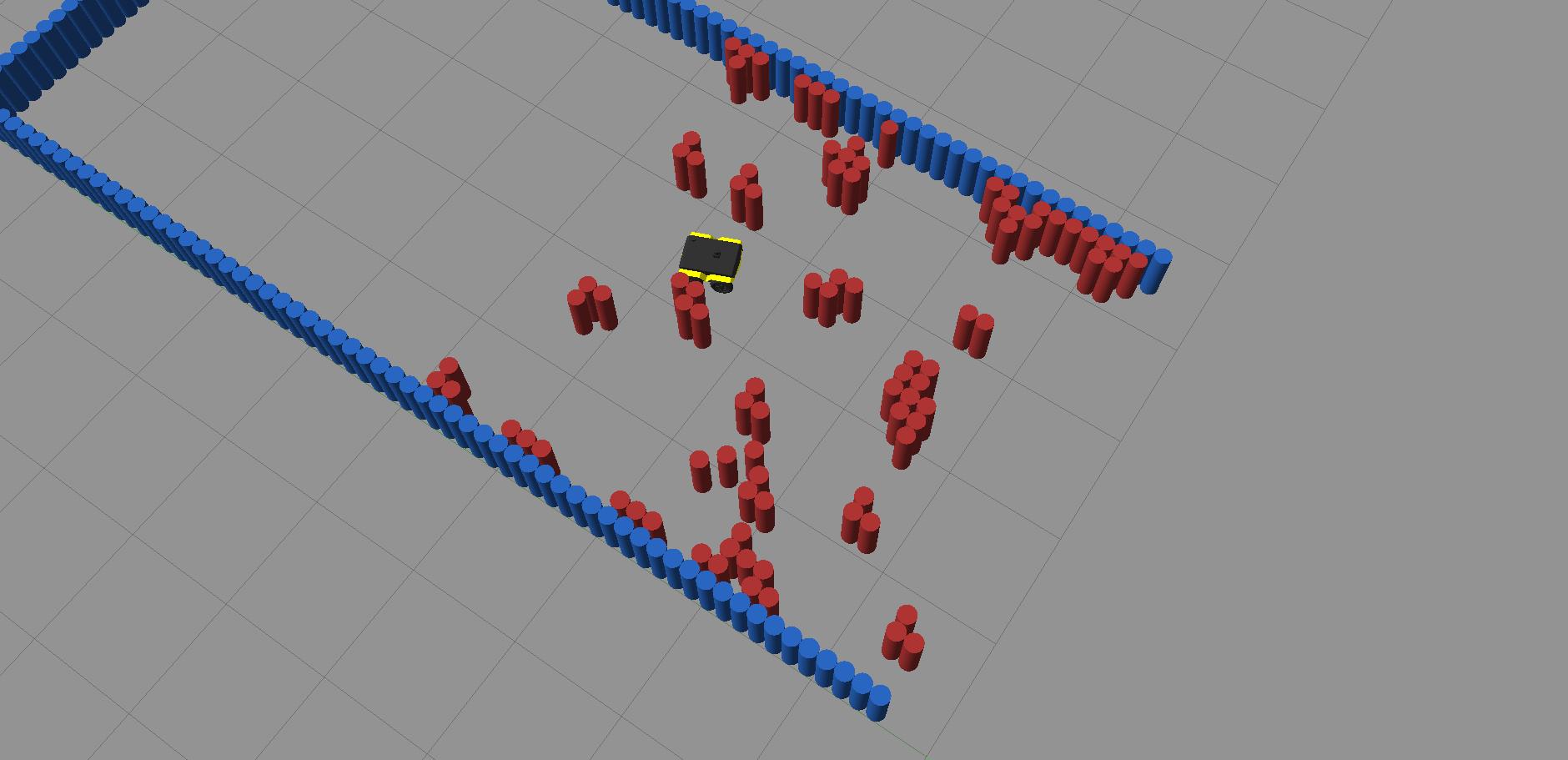} }}%
            \quad
            \subfloat[\centering World Index 134]{{\includegraphics[trim={15cm 5cm 15cm 0cm},clip=true, width=5cm]{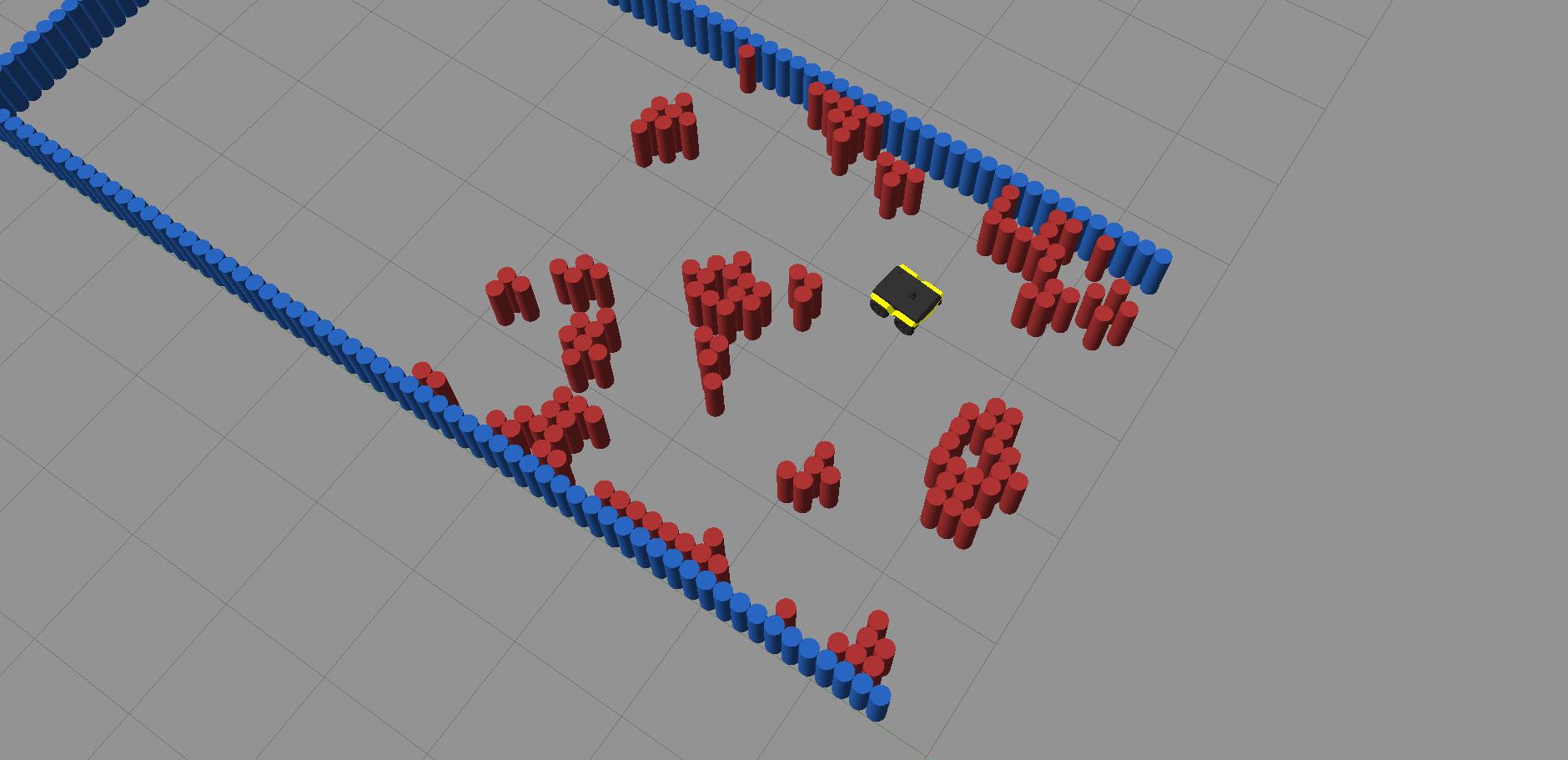} }}%
            \quad
            \subfloat[\centering World Index 288]{{\includegraphics[trim={15cm 5cm 15cm 0cm},clip=true, width=5cm]{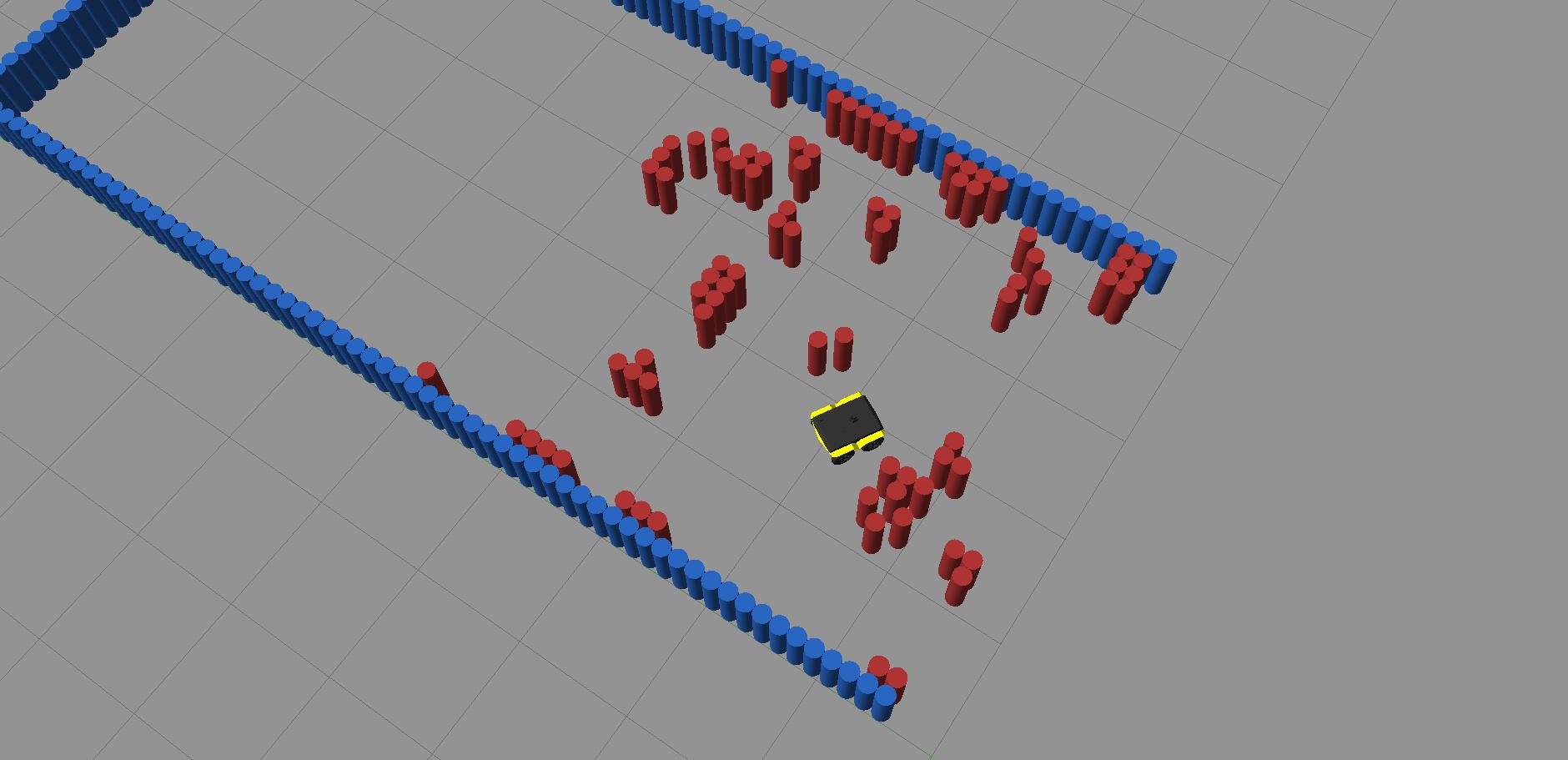} }}%
            \quad
            \caption{The BARN challenges simulation environments on a different difficulty.}%
            \label{fig:simulation_environment}%
        \end{figure}

        The BARN Challenge simulation environment \cite{perille2020benchmarking} consists of 300 procedural-generated scenarios. The difficulty is parameterized to range from easy, with a lot of open space, to hard, with tight spaces. Each scenario is generated via a combination of three algorithms: a cellular automaton algorithm \cite{wolfram1983statistical}, a flood-fill algorithm \cite{torbert2016applied}, and the A\textsc{*} path-finder algorithm \cite{hart1968formal}. First, the cellular automaton produces a binary grid to represent obstacles and free spaces. The resulting configuration is conditioned on 4 parameters: {initial fill percentage, number of smoothing iterations, fill threshold, and clear threshold}. Next, the flood-fill algorithm validates the existence of a pathway between generated points. Finally, A\textsc{*} algorithm is utilized to plan a feasible path from the start to the goal position. If there is no feasible path, then the map is discarded. Also, the length of this path is used to determine the difficulty of the generated scenario (Eq.~\ref{eq:optimal_traversal}). Examples of the generated simulation scenarios are shown in Fig.~\ref{fig:simulation_environment}.
        
        \par The BARN environment generator \cite{perille2020benchmarking} can be reproduced using the provided source code available at the following repository \footnote{\url{https://github.com/dperille/jackal-map-creation}}.

    \subsection{Evaluation}
        \par Each team submits its navigation system according to a standardized pipeline \footnote{\url{https://github.com/Daffan/nav-competition-icra2022}} provided by the organizers. The submissions are executed for 10 trials on all 300 simulation scenarios. The final submission score is computed by combining the success rate (reaching the goal without any collisions), average traversal time, and the difficulty of the scenario (computed as the optimal traversal time in Eq. \ref{eq:optimal_traversal}). The score $s_i$  of each environment is computed from the following expression:
        
        \begin{equation} 
            \label{eq:sucess_rate}
            s_i = 1^{\textrm{success}}_i \times \frac{\textrm{OT}_i}{\textrm{clip}(\textrm{AT}_i, 4\textrm{OT}_i, 8\textrm{OT}_i)} \; ,
        \end{equation}
        
        \noindent where the $1_{success}$ indicates whether the robot successfully $(1)$ reached the goal in the given time, or collided with an obstacle or timed out $(0)$. $AT$ denotes the actual traversal time and $OT$ indicates the optimal traversal time based on the environment's difficulty assuming the robot travel at the maximal speed of $2$ m/s and followed the path computed by the A\textsc{*}. $OT$ is computed according to the following equation:
       
        \begin{equation} 
            \label{eq:optimal_traversal}
            \textrm{OT}_i = \frac{\textrm{Path Length}_i}{\textrm{Maximal Speed}}.
        \end{equation}

        \par In the real-world stage, the evaluation metric was simply success rate and actual traversal time. If multiple teams achieved the same success rate (tie), the team with the fastest time wins the competition.
        
\section{Our Approach}
    \par Our base approach was to extend the best-performing baseline of Learning from Learned Hallucination (\textit{LfLH}) \cite{wang2021agile} with improved collision check and recovery behaviors via finite-state machines (FSM). We introduce two alternatives for forward safety checks: footprint inflation (FI) and model predictive control (MPC). The approach relies on a learning-based model that learns to drive the robot by collecting random trajectories and hallucinating obstacles \cite{xiao2021toward, wang2021agile}. In our early experiments, we found that the original formulation of the FSM method for recovery tended to get stuck and even collide in some scenarios.
    
    \par In this section, we describe our navigation solution to control the Jackal robot. Starting from the self-designed FSM, followed by backtracking recovery behavior, and finally, the different approaches used in the simulation and in the real-world stage. 

    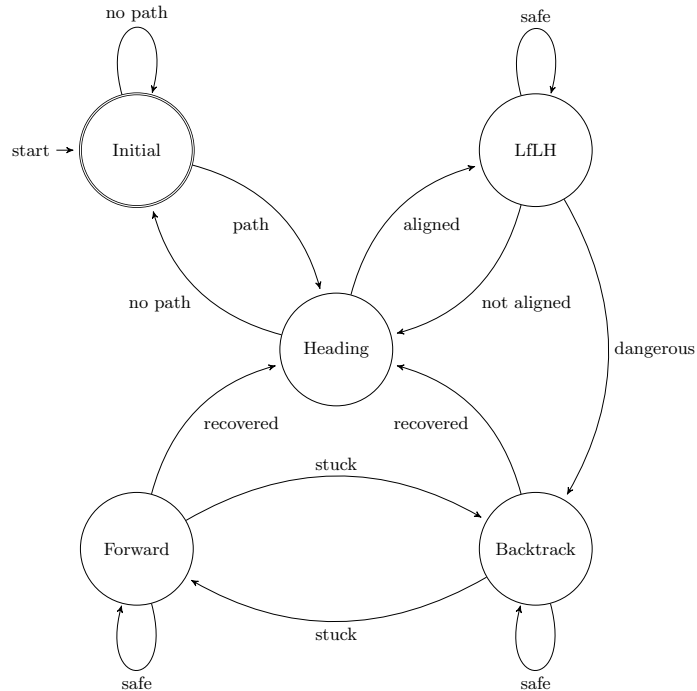
\begin{figure}
        \centering
        \scalebox{0.75}{\usetikzlibrary{arrows, automata}

\begin{tikzpicture}[>=stealth', shorten >=3pt, auto, node distance=5cm]
     \node[initial,state, circle, accepting, minimum size=2.0cm]       (initial)     {Initial};
     \node[state, circle, minimum size=2.0cm]  (heading)   [below right of=initial]  {Heading};
     \node[state, circle, minimum size=2.0cm]  (LfLH)      [above right of=heading]  {LfLH};
     \node[state, circle, minimum size=2.0cm]  (backtrack) [below right of=heading]  {Backtrack};
     \node[state, circle, minimum size=2.0cm]  (forward)   [below left of=heading]   {Forward};

     \path[->] (initial)   edge [loop above]   node {no path}       (initial)
               (initial)   edge [bend left]    node[below left] {path}          (heading)
               (heading)   edge [bend left]    node {no path}       (initial)
               (heading)   edge [bend left]    node[below right] {aligned}       (LfLH)
               (LfLH)      edge [loop above]   node {safe}          (LfLH)
               (LfLH)      edge [bend left]    node {not aligned}   (heading)
               (LfLH)      edge [bend left]    node {dangerous}     (backtrack)
               (backtrack) edge [loop below]   node {safe}          (backtrack)
               (backtrack) edge [bend left]    node {stuck}         (forward)
               (backtrack) edge [bend right]   node {recovered}     (heading)
               (forward)   edge [loop below]   node {safe}          (forward)
               (forward)   edge [bend left]    node [below right] {recovered}     (heading)
               (forward)   edge [bend left]    node {stuck}         (backtrack);
\end{tikzpicture}}
        \caption{Finite state machine of Inventec team.}
        \label{fig::fsm}
    \end{figure}
    
    \subsection{Finite State Machine}
    \par We designed a navigation FSM to handle the state behavior in our local controller. It includes five states: \textit{Initial, Heading, LfLH, Forward, and Backtrack}, as shown in Fig. \ref{fig::fsm}. In the \textit{Initial} state, the navigation controller waits for the path computed with Dijkstra’s search in the \textit{move\_base} global planner with \textit{NavFn} plugin. Then, the state is switched to \textit{Heading}, which aligns the robot to the target path. The target is to maintain the robot's alignment within a tolerance of $\pm 30^{\circ}$. \textit{Heading} is triggered whenever heading deviation is outside the desired tolerance.

    \par When the heading is aligned, we employ the \textit{LfLH} model to produce velocity commands from the input of LiDAR and a position sampled from the path 0.5m ahead of the robot. This is the main driving mode, and our goal is to stay in this state as long as possible. However, the commands are not perfect, so we run a forward safety check at every step. If a commanded velocity is considered dangerous, i.e. results in a future collision, the state is changed to \textit{Backtrack} recovery behavior. In \textit{Backtrack}, we applied a backtrack safety check for the backtrack movement using Region-of-Interest (ROI) in the costmap. In cases where the backtracking movement poses a risk by moving too close to previously seen obstacles, the state transitions to executing slow \textit{Forward} movement while monitoring the robot's recovery condition.
    
    \subsection{Recovery Behavior}
    
    \par
\begin{figure}
        \centering
        \subfloat[\centering LiDAR field-of-view in Jackal robot.] {\label{fig::lidar_fov}{\includegraphics[width=0.3\columnwidth]{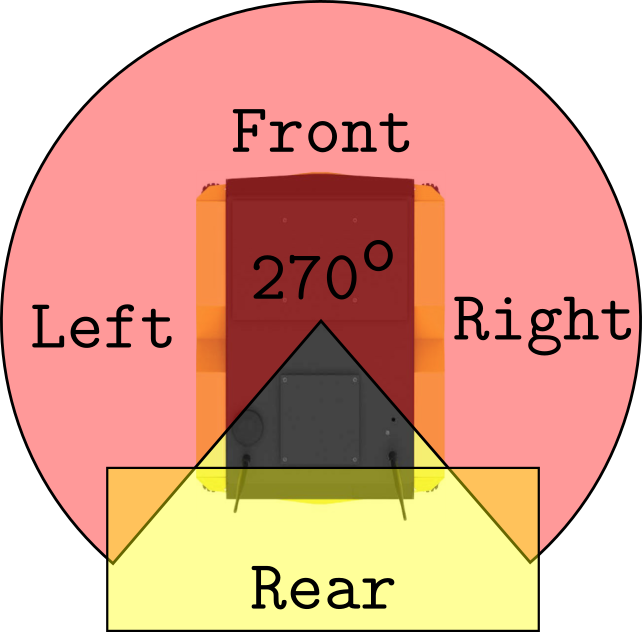} }}
        \hspace{10mm}
        \subfloat[\centering The Costmap with robot rear ROI (yellow rectangle)]{{\label{fig::jackal_costmap}\includegraphics[width=0.3\columnwidth]{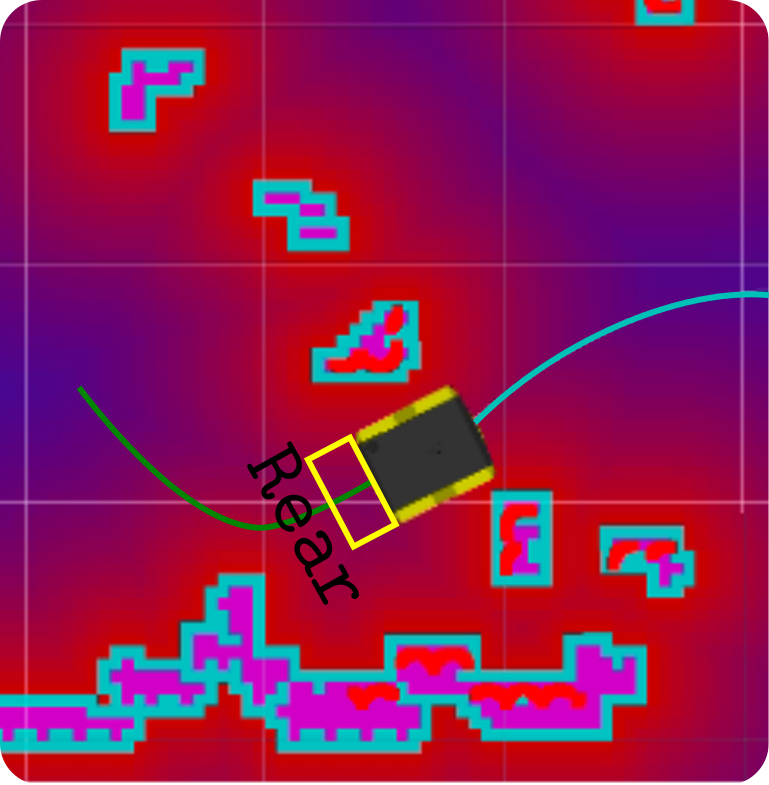} }}
        \caption{The Jackal LiDAR and costmap.}
        \label{fig::jackal_lidar_costmap}
    \end{figure}

 Even a strong controller can result in imperfect commands due to model inaccuracies or unexpected obstacles. To recover under such circumstances, we implemented recovery behaviors in the form of backtracking and slow-forward commands. The strategy for backtracking consists of recording the robot's path (green line shown in Fig.\ref{fig::jackal_costmap}) during a forward movement. When \textit{Backtrack} is first triggered, it samples a point 0.3 meters behind the robot in the recorded path. Then, the robot will first align to the target point and then perform a straight backward command. We compute the Euclidean distance to monitor the robot's arrival at the designated position continuously.

    \par Moving backward means we're moving towards the blind spot of the LiDAR, and potentially into obstacles the robot can't detect. To address the limitation of LiDAR coverage, we define a rectangular region of interest (ROI) directly behind the robot, illustrated in Fig.~\ref{fig::lidar_fov}). At every step, we check the ROI with past costmap information to detect possible collisions during recovery. The state is switched over to slow-forward recovery if a potential collision is detected during the reverse movement.

    \par A slow-forward movement with $0.2$ m/s effectively solved cases where the robot got stuck in some scenarios. Because the global planner continuously recalculates the path to reach the goal. It prevents the robot from becoming stranded in one location. However, in some cases where the same path is given, it may get trapped in an infinite loop of moving forward and backward within a specific area.

    \subsection{Simulation Approaches}
        \par In the simulation stage, we utilized the LfLH model from \cite{wang2021agile} as the main controller in our navigation system. In \cite{wang2021agile}, they proposed two models with different max velocities: $1$ m/s and $2$ m/s. Notably, both models are trained with the same model configuration. Subsequently, we compared both models' performance in the 300 standard BARN environment and observed the slower model outperformed the faster model achieving scores of 0.2304 ad 0.2158 respectively.


        \begin{figure}
            \centering
            \includegraphics[width=0.7\columnwidth]{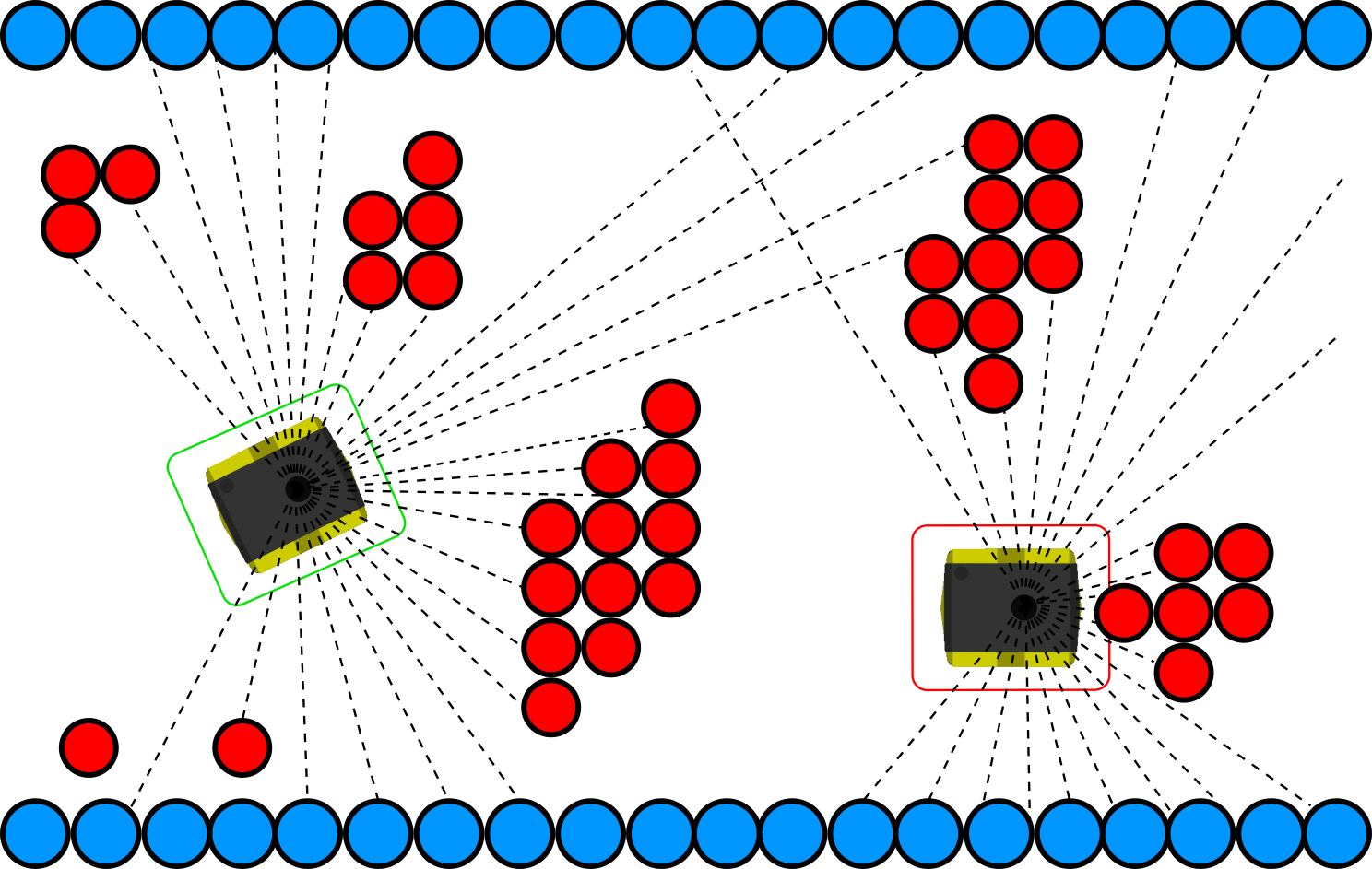}
            \caption{The footprint inflation detection.}
            \label{fig::jackal_inflation}
        \end{figure}
        
        \par Additionally, we conducted retraining of the model to compare its performance with the pre-trained model available in their public repository \footnote{\url{https://github.com/LearningUAV/hallucination}}. As a result, due to minimal variation in performance, we opted to utilize the original, readily available $1$ m/s LfLH model. Also, based on the comparative analysis of baseline performance, we identified the primary reason for the failure in the fastest model is attributed to the slippage when braking at high-speed.
        
        \par Our main strategies for improving the baseline model are threefold. First, we add footprint inflation for the additional obstacle check based on the real-time LiDAR data (shown in Fig. \ref{fig::jackal_inflation}). Then, we check the costmap for historical obstacles to assess the safety of the robot's rear side. Finally, we clip the maximum velocity to $0.7$ m/s, which also provides a significant performance improvement.
        
        \par Footprint Inflation (FI) effectively checks the inclusion of LiDAR points within the rectangle boundary. We applied a $0.04$ m offset from the actual dimension of the robot, which is $0.43$ m x $0.508$ m (width x length). As a result, whenever the inflated footprint contains LiDAR detected obstacles, the robot will stop and transition into recovery behavior. Two illustrations of the inflated footprint are depicted in Fig. \ref{fig::jackal_inflation}, where the green region and red region indicate safe and unsafe conditions respectively. Moreover, we utilized the \textit{mpl\_path} \footnote{\url{https://matplotlib.org/stable/api/path_api.html}} library from matplotlib for the real-time boundary contain points detection.

        \par During the benchmark of BARN simulation environment, this method enables the robot to navigate in close proximity to obstacles without collision. However, there exists a trade-off between the size of the inflated footprint and the maximum velocity. If the size is too large, the robot will have enough time for braking at high speed but is unable to navigate in a highly constrained environment as it frequently stops. Conversely, if the boundary size is too small, there is a higher probability of front collision due to insufficient time for braking. We conducted a test and determined that a $0.7$ m/s  maximum velocity paired with a $0.04m$ offset provided the optimal balance. This configuration allows the robot to stably navigate without encountering a front collision.

        \begin{figure}
            \centering
            \includegraphics[width=0.7\columnwidth]{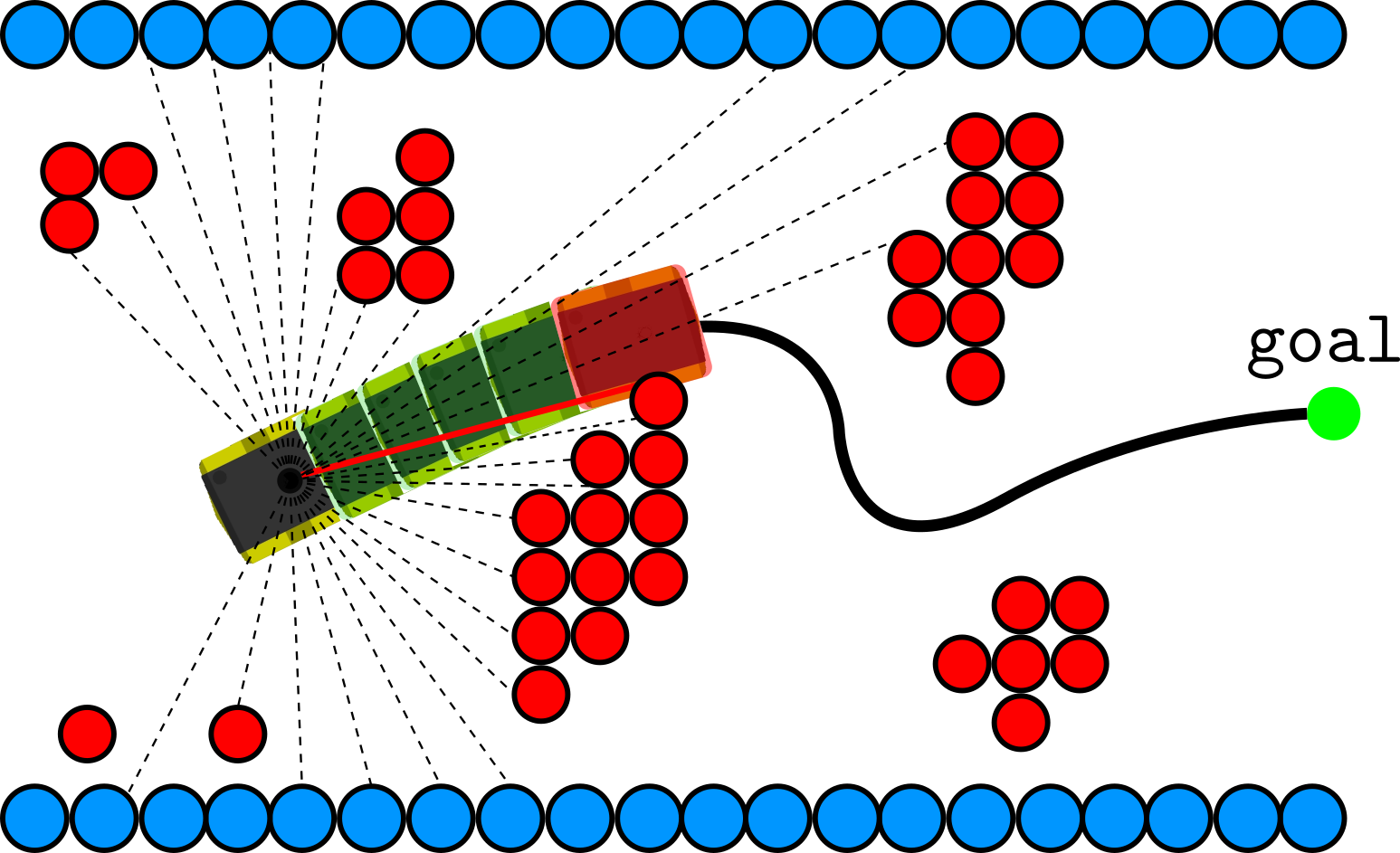}
            \caption{The MPC footprint forward safety check based on LiDAR sensor.}
            \label{fig::real_mpc}
        \end{figure}

    \subsection{Real-World Approaches}
        \par Our approach in the real-world is similar to the simulation but with a different method for determining safety when moving forward. Instead of using the footprint inflation method from the simulation, we opted for a Model Predictive Control (MPC) approach.
        
        \par During the \textit{LfLH} operation we run an MPC to predict the future steps and detect collisions. At every step, we predict 20 steps into the future (about 200 milliseconds). If at any point in the future trajectory, an obstacle overlaps with the footprint of the robot, the current command is deemed unsafe. This stops the \textit{LfLH} and the FSM switches to recovery behavior. Fig.~\ref{fig::real_mpc} depicts this process, with green indicating safe portions of the trajectory and red showing a detected future collision.

\section{Results and Discussion}
    \par In this part we deliberate our results and discuss the lessons that we learned from BARN 2023. First, in the results, we break our approach into two parts: the simulation and physical competition. Then, we discussed what can be improved and the future application related to this challenge.   
    
    \subsection{Simulation}
        \par Both of our FI and MPC approaches managed to beat the \textit{LfLH} baseline \cite{wang2021agile}. The MPC submission achieved a score of $0.2419$, an improvement of $3.64$\% over the baseline. Our best-performing approach was the FI approach with a score of $0.2445$ and ranking 2nd place overall. This is the score shown on the final simulation leaderboard as seen in Tab.~\ref{tab::sim_results}.

        \begin{table}
          \caption{Simulation Results.}
          \label{tab::sim_results}
          \centering
          \small
          \begin{tabular}{c|c|c}
          \toprule
          Rank. & Team/Method (University) & Score \\
          \midrule
          1 & KUL+FM & 0.2490\\
          2 & INVENTEC (Inventec Corp) & 0.2445\\
          3 & University of Almeria & 0.2439\\
          4 & UT AMRL (The University of Texas at Austin) & 0.2424\\
          5 & LfLH (Baseline~\cite{wang2021agile}) & 0.2334\\
          6 & TEMPLE TRAIL (Temple University) & 0.2290\\
          7 & UVA AMR (University of Virginia) & 0.2237\\
          8 & RIL (Indian Institute of Science, Bengaluru) & 0.2203\\
          9 & E-Band (Baseline~\cite{quinlan1993elastic}) & 0.2053\\
          10 & End-to-End (Baseline~\cite{pfeiffer2017perception}) & 0.2042\\
          11 & Staxel & 0.2019\\
          12 & APPLR-DWA (Baseline~\cite{xu2021applr}) & 0.1979\\
          13 & The MECO Barners (KU Leuven) & 0.1829\\
          14 & Fast ($2.0$m/s) DWA (Baseline~\cite{fox1997dynamic}) & 0.1709\\
          15 & Default ($0.5$m/s) DWA (Baseline~\cite{fox1997dynamic}) & 0.1627\\
          \bottomrule
          \end{tabular}
        \end{table}

        \par In our experiments, we found that footprint inflation outperformed MPC in certain scenarios due to its ability to provide a more conservative safety margin, ensuring a higher degree of collision avoidance. The offset variance of footprint inflation was indeed a concern, as it could lead to finding unbalanced pair parameters with maximum linear velocity. Resulting, in unpredictable behaviors and potentially compromising the robot's safety in real-world environments.

        \par The decision to use MPC in the real-world, despite the success of FI, was driven by the need for a more dynamic and adaptive approach. MPC takes into account not only the robot's current footprint but also integrates future position information, enabling the robot to proactively plan its movement and adjust its trajectory accordingly. This reinforced its suitability for predictive collision, resulting in adaptability and real-time responsiveness in real-world navigation.
        
    \subsection{Real-World Stage -- London}
        \par
\begin{figure}
            \centering
            \includegraphics[trim={3cm 3cm 3cm 3cm},clip=true, width=0.6\columnwidth]{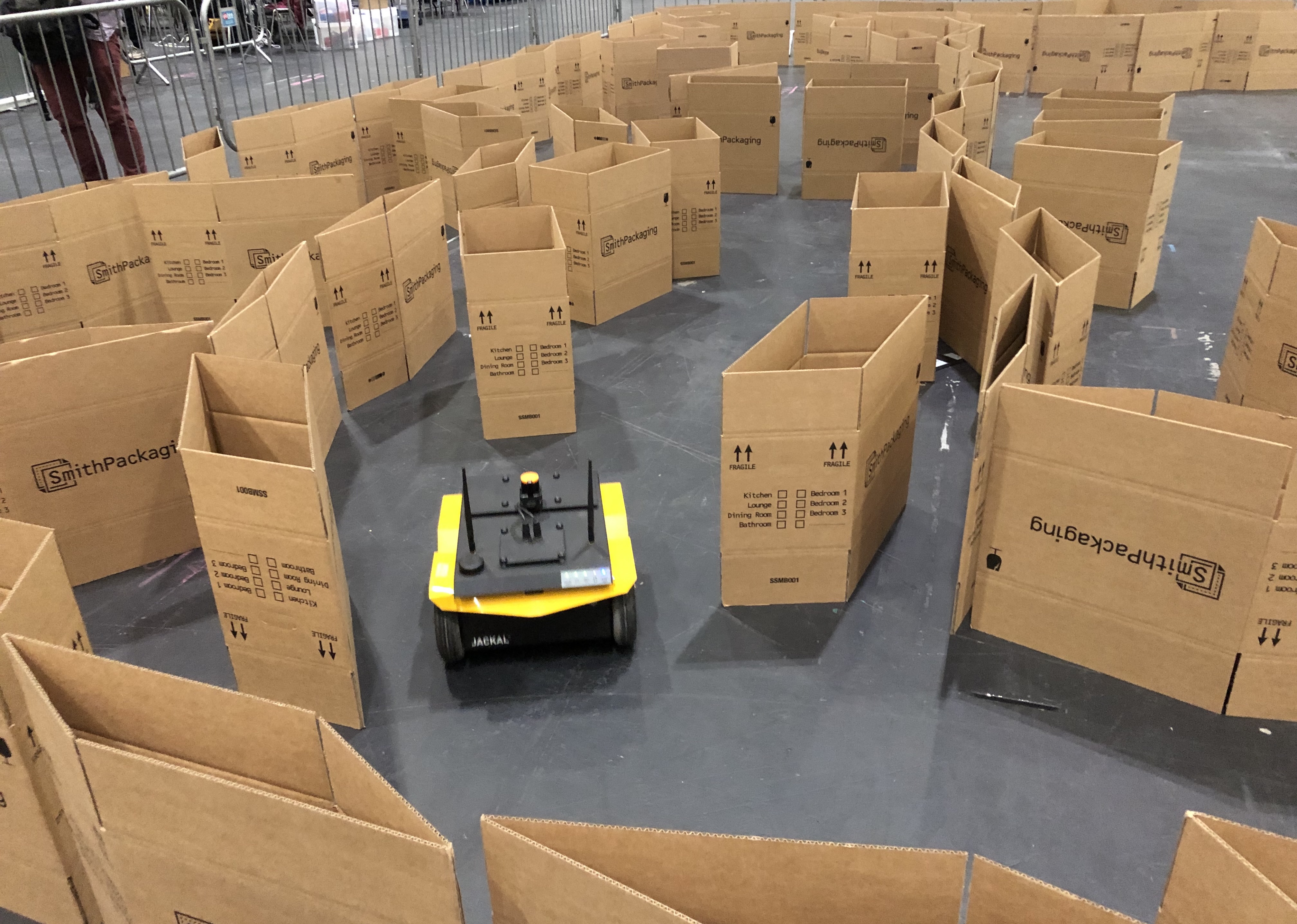}
            \caption{The BARN physical challenge in ICRA 2023, London.}
            \label{fig::barn_physical}
        \end{figure}

 Due to ranking 2nd place on the simulation leader board, we were invited to participate in the world stage during the ICRA 2023 event in London. The physical BARN challenge was joined by the top participants to showcase their performance in the standardized Jackal robot (refer Fig. \ref{fig::barn_physical}). During this event, the organizer set up three different courses to benchmark each team's performance.

        \par The three scenarios were designed to gradually increase in difficulty based on team performance. In the first round, the obstacles were set up to be "easy", offering a manageable challenge for all participating teams. However, as teams demonstrated their proficiency and successful navigation in the initial rounds, the subsequent rounds featured progressively harder obstacles. This approach allowed for fair competition and encouraged teams to continuously improve their robot's capabilities and adapt to increasingly complex environments.

        \begin{table}
            \caption{Physical Results.}
            \label{tab::physical_results}
            \centering
            \small
            \begin{tabular}{c|c|c}
            \toprule
            Rank. & Team/Method (University) & Success / Total Trials \\
            \midrule
            1 & KUL + FM & 9/9\\
            2 & INVENTEC & 6/9\\
            3 & University of Almeria & 5/9\\
            \bottomrule
            \end{tabular}
        \end{table}

        \par Our team successfully completed six out of the nine challenging courses, ranking in 2nd place. Since no teams were tied, the final ranking was solely based on the number of successful course completions and shown in Tab.~\ref{tab::physical_results}. Our approach was robust in the first and second courses, smoothly navigating in close proximity to obstacles. It also prevented the robot from executing unsafe high-velocity commands in constrained spaces. However, our solution was proven sub-optimal in the last round, with the hardest obstacles. We had issues with our planner in open spaces, that were different than what we tuned for in simulation. Also, our approach was too conservative and caused the robot to get stuck in sections going back and forth.

    \subsection{Discussion}        
        \par In addition to having a low-level controller that produces good-quality velocity commands, tuning of the path planner in the navigation system plays an important role in successful navigation. Our approach did not use any mapping, which caused the robot to get stuck in sections of the course with open areas. In the future, we will implement a SLAM approach that ensures the robot only explores ``forward'' in the course. This should minimize the need for frequent re-planning and getting stuck.

\section{Conclusion}
    \par
\begin{figure}
        \centering
        \includegraphics[trim={2cm 3cm 2cm 8cm},clip=true, width=0.85\columnwidth]{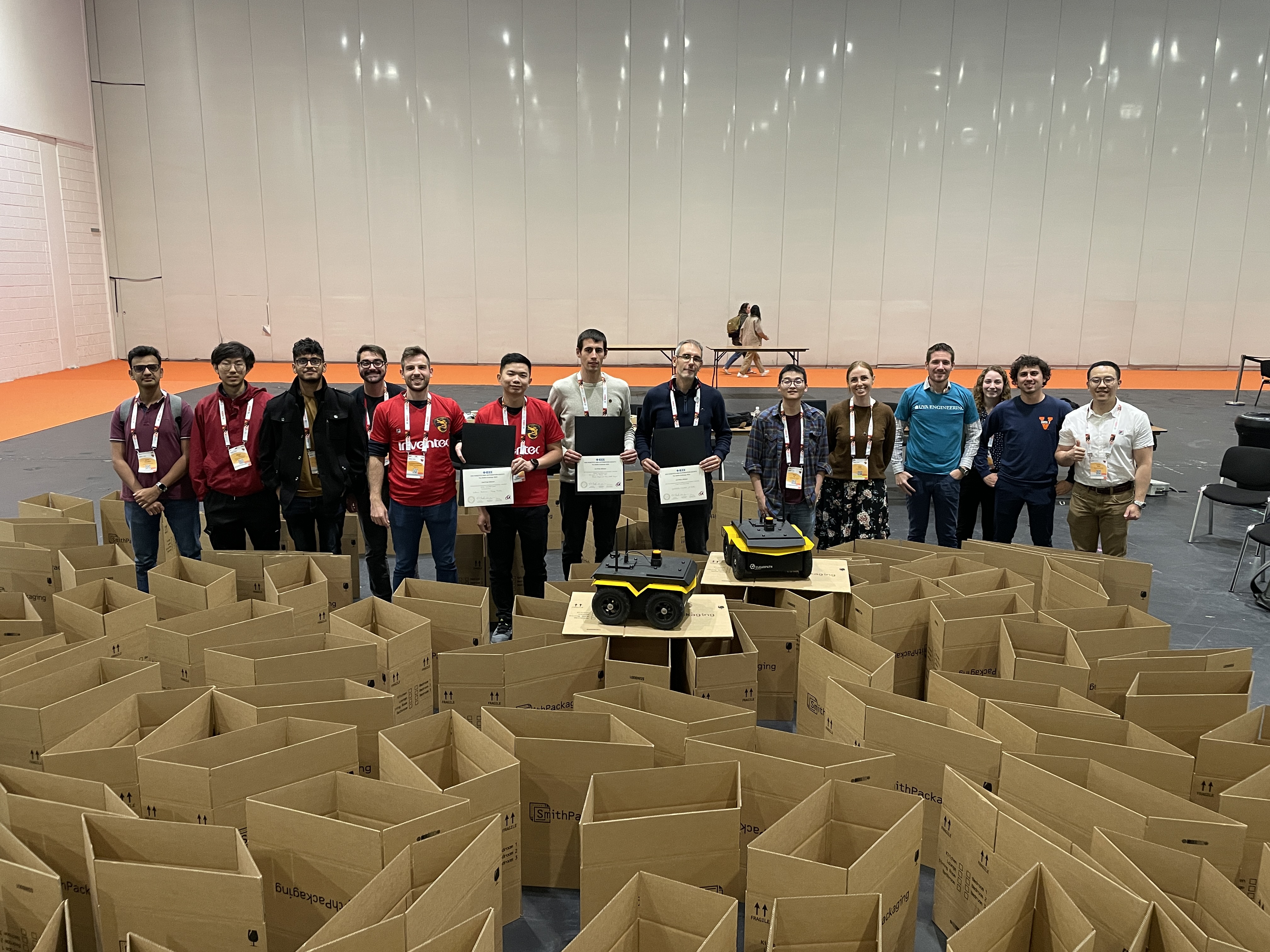}
        \caption{The closing ceremony of BARN challenge 2023.}
        \label{fig::closing_ceremony}
    \end{figure}

 This report described the participation of the Inventec team BARN Challenge 2023. The BARN challenge consists of a navigation challenge in highly constrained spaces. We showcased our proposed navigation system capabilities in both simulation and real-world environments. Our approach relied on extending the baseline LfLH \cite{wang2021agile} performance, with a score improvement from $0.2334$ to $0.2445$ ($4.76\%$) in the simulation environment. Overall, we ranked 2nd place in both simulation and physical competition (refer to Fig. \ref{fig::closing_ceremony}).

    \par Our approach utilized a finite state machine to manage moving forward with a learning-based controller and recovery behaviors. We discussed two alternatives for front safety collision checks: Footprint Inflation (used in the simulation stage) and Model Predictive Control (used in the real-world stage). Also, the backtrack safety check, based on the Region-of-Interest in costmap obstacle history information.

    \par However, we also encountered some challenges, particularly when transitioning from the simulation environment to the real-world, generalizing path planning in both tight and open spaces. In the future, we wish to address these shortcomings and improve the robustness and versatility of robot navigation systems for future competition and other real-world scenarios.

\section{Acknowledgment}
    \par We would like to express our gratitude to Wei-Chao Chen (Chief Digital Officer and Senior Vice President of Inventec Corporation) for his guidance, support, and insightful feedback throughout this project. We are truly grateful for the opportunities provided, the trust placed in us, and the encouragement that has propelled us forward.
    \par We also extend our appreciation to the competition organizers, Xuesu Xiao and Zifan Xu, as well as Clearpath Robotics, for their work during The BARN Challenge 2023. Their dedication, organization, and attention to detail resulted in a smooth competition. Their contributions have brought innovation and promoted collaboration in advancing robot navigation technologies. Despite the competitive nature of the event, every participating team was friendly, open to discussion, and made for a great time.

\bibliographystyle{IEEEtran}
\bibliography{IEEEabrv,references}
\end{document}